\documentclass[letterpaper, 10 pt, conference]{ieeeconf}

\IEEEoverridecommandlockouts
\usepackage{algorithm}
\usepackage{amsmath}
\usepackage{amsfonts}
\usepackage{amssymb}
\usepackage{algpseudocode}
\usepackage[caption=false,font=footnotesize]{subfig}
\usepackage{adjustbox}
\usepackage{comment}
\usepackage{xcolor}
\usepackage{newfloat}
\usepackage{listings}
\usepackage{soul}
\usepackage{pifont}
\usepackage{multirow}
\usepackage{booktabs}
\usepackage{hyperref}
\usepackage{booktabs,tabularx,array,makecell,graphicx}
\definecolor{lightgray}{gray}{0.85}
\newcommand{\bbZ}{\mathbb{Z}}
\newcommand{\cmark}{\ding{51}}
\newcommand{\xmark}{\ding{55}}
\newcommand{\dist}{d}
\renewcommand{\st}[1]{\textsf{#1}}
\newcolumntype{Y}{>{\centering\arraybackslash}X}
\newcolumntype{L}[1]{>{\raggedright\arraybackslash}p{#1}}
\title{Lifelong Multi-Subsystem Pickup and Delivery\\
with Buffer-Limited Handover Stations}

\author{Chuanlong Zang$^{1,2}$, Isabelle Barz$^{1}$, Anna Mannucci$^{1}$, Philipp Schillinger$^{1}$, Florian Lier$^{1}$, Wolfgang Hönig$^{2,3}$
\thanks{$^{1}$Robert Bosch GmbH, Corporate Research, Stuttgart, Germany.
{\tt\small Chuanlong.Zang@de.bosch.com}}%
\thanks{$^{2}$Technical University of Berlin, Berlin, Germany.}%
\thanks{$^{3}$Robotics Institute Germany (RIG).}%
}

\begin{document}
\maketitle
\thispagestyle{empty}
\pagestyle{empty}
\bstctlcite{mybstctl} 

\begin{abstract}
Coordinating payload transfers between subsystems is a critical challenge in lifelong Multi-Agent Pickup and Delivery (MAPD). We study systems where agents are confined to separate regions and must exchange payloads through shared handover stations. These stations, equipped with single docks and finite buffers, are inherently vulnerable to blocking and starvation. We formalize this problem as Multi-Subsystem MAPD with Buffer-limited Handover Stations (MS-MAPD-BHS). We then propose Handover-Aware Reservation and Routing (HARR), an online controller that couples per-subsystem planners. HARR uses a shared dock reservation calendar and a deterministic rolling-horizon projection of buffer occupancy to coordinate actions. A candidate route is accepted only if its dock interval is free and the resulting buffer occupancy projection remains within capacity. Under perfect execution, these checks ensure collision-free dock use and buffer-safe committed operations within the reservation horizon. In simulation, HARR achieves up to 77\% higher throughput and 92\% lower backlog than a fixed-dock ablation at moderate load, while also reducing planning time relative to a coupled station-aware Token Passing baseline. These results show that explicit interface coordination substantially improves stability in modular multi-subsystem transport.
\end{abstract}
\section{Introduction}
Multi-Agent Pathfinding (MAPF) and especially its lifelong variants, such as Multi-Agent Pickup and Delivery (MAPD), provide foundational tools for coordinated transport in warehouses, factories, and traffic systems. These frameworks couple task allocation with collision-free routing under continuous task arrivals~\cite{li_scalable_2021, ma_lifelong_2017, he_scaling_2024}. While powerful, they typically assume a single, globally-coordinated fleet.

However, many real-world deployments are inherently \emph{multi-subsystem}, where distinct agent fleets operate within separate regions --- often under different management --- but must collaboratively exchange payloads at shared interfaces to complete end-to-end transport. Examples range from multi-vendor logistics, robotic systems interfacing with conveyors to heterogeneous cooperative teams and interacting subsystems in terminals~\cite{chen_yard_2020, fragapane_planning_2021, menebroker_mobile_2025, arbanas_decentralized_2018}. The primary challenge in such distributed architectures lies in coordinating the \emph{interfaces} between subsystems without sacrificing modularity or scalability.

Our focus is on the common scenario where subsystems exchange payloads exclusively through fixed handover stations. Unlike in standard MAPD, each handover station is a critical shared resource: it has one dock vertex with exclusive access, non-preemptive operation times, and a finite-capacity buffer. Under realistic load, these constraints can cause upstream \emph{blocking} (an unload cannot start because the buffer is full) and downstream \emph{starvation} (a load cannot start because the requested payload is unavailable), even when motion within each individual subsystem is efficient~\cite{bonetti_agv_2024,chung_deadlock_2024}. Here, \emph{buffer safety} means that no accepted operation causes a station buffer to underflow or exceed its capacity, whereas \emph{availability coordination} means that a downstream load is reserved only for a payload already present at that station. Consequently, merely running independent MAPD solvers and allowing agents to `queue' at the interface can commit mutually incompatible station operations.

To address this challenge, we formalize \emph{MS-MAPD-BHS} (\textbf{M}ulti-\textbf{S}ubsystem MAPD with \textbf{B}uffer-limited \textbf{H}andover \textbf{S}tations) and propose \emph{HARR} (Handover-Aware Reservation and Routing). HARR schedules two-sided station operations while respecting dock exclusivity and finite-buffer dynamics. Its interface state is a deterministic schedule derived from operations that have already been accepted.

\begin{figure}[t]
\centering
\includegraphics[width=\columnwidth]{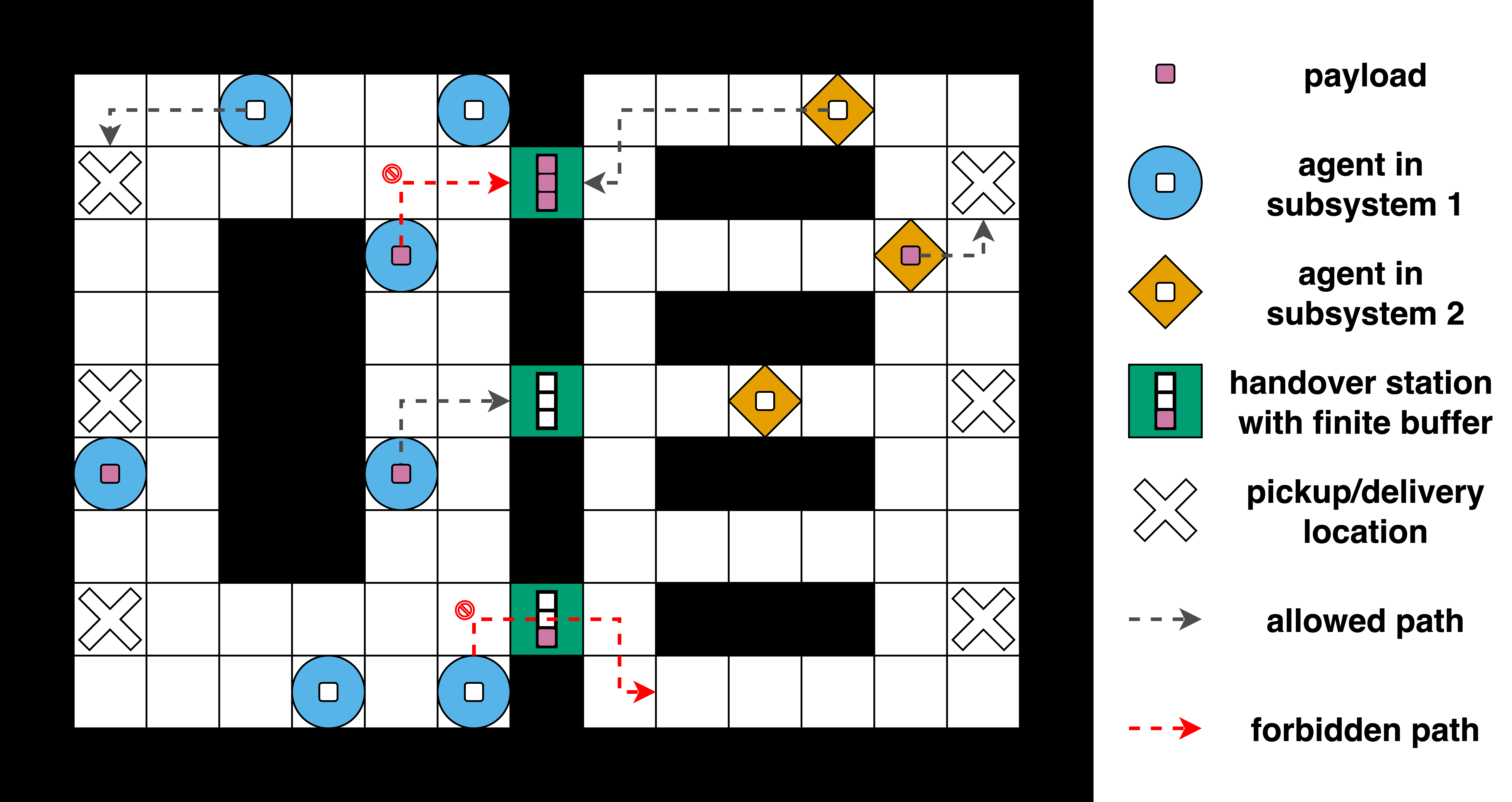}
\caption{Two-fleet lifelong MAPD with buffer-limited handover stations. The crossed-circle marking on the upper route denotes a dock entry that is inadmissible \emph{at the illustrated timestep} because the loaded upstream agent would unload into a full buffer; it does not make the entire route permanently forbidden. The agent may wait or replan and enter after capacity becomes available.}
\label{fig:system_overview}
\end{figure}

The main contributions of this paper are:
\begin{itemize}
    \item We formally define \emph{MS-MAPD-BHS}, a lifelong multi-subsystem pickup-and-delivery problem where agents are confined to regions and inter-subsystem transfers occur only through shared handover stations with single-dock access, non-preemptive load/unload times, and finite-capacity buffers with explicit dynamics (Fig.~\ref{fig:system_overview}).
    \item We propose \emph{HARR}, an online controller that couples per-subsystem Token Passing (TP) planners \cite{ma_lifelong_2017} using a shared dock calendar and a deterministic rolling-horizon buffer occupancy projection. Candidate plans are accepted only when the dock interval is free and the resulting committed buffer schedule remains feasible.
    \item We evaluate mean service time under online task arrivals, comparing HARR against a coupled TP reference and a fixed-dock ablation, measuring service time, throughput, backlog, and planning cost.
\end{itemize}
\section{Related Work}
\label{sec:related}

MAPF methods range from optimal search variants to bounded-suboptimal, large-neighborhood, prioritized, and rolling-horizon approaches~\cite{stern_multi-agent_2019,gao_review_2024}. Package-exchange and shared-infrastructure variants also exist~\cite{ma_multi-agent_2016,he_conflict-based_2026}, but do not address lifelong finite-buffer transfer.

Online Multi-Agent Pickup and Delivery (MAPD), where tasks arrive dynamically, is commonly addressed via Token Passing (TP) and reservation-table methods~\cite{ma_lifelong_2017, ma_lifelong_2019}. Other MAPD variants incorporate richer objectives and constraints like task deadlines ~\cite{makino_online_2024, huang_distributed_2026}, external agents \cite{bonalumi_multi-agent_2025}, kinematic constraints~\cite{ma_lifelong_2019}, or exchange coordination (e.g., PERR, MAPD-MP~\cite{flammini_multi-agent_2025}, MAPF-E~\cite{he_conflict-based_2026}, MAPD-HA~\cite{flammini_multi_2026}). These formulations enrich tasks and motion, but do not model a finite-capacity, single-server transfer interface that stores payloads between independently operating fleets.

Cooperation among heterogeneous teams is also studied outside MAPD. For example, decentralized hierarchical planning coordinates aerial and ground vehicles with distinct capabilities~\cite{arbanas_decentralized_2018}. Such work motivates inter-team task decomposition and coordination, but does not address lifelong collision-free payload flow through a shared finite buffer. Conversely, parcel-transshipment research studies terminal capacity, sorting resources, blocking, and combined simulation and optimization~\cite{mcwilliams_transshipment_2014,clausen_transshipment_2017}. These models provide relevant operational context for handover bottlenecks, but typically aggregate parcel flow rather than plan time-indexed trajectories for individual robots.

Table~\ref{tab:mapd_variant_comp} positions MS-MAPD-BHS along five dimensions: online task arrivals, multiple fleets or environments, explicit payload transfer, finite transfer-point storage, and time windows or deadlines. Existing work covers several of these dimensions separately, but not their combination. In particular, it does not explicitly model the buffer dynamics that create blocking and starvation at an asynchronous handover interface, as highlighted by the \textbf{Buf} column.

HARR is complementary to work on joint task allocation and path planning~\cite{honig_conflict-based_2018,fan_joint_2025,chen_integrated_2021}. Rather than redesigning allocation or MAPF solvers, it augments scalable online MAPD with explicit dock and buffer calendars while keeping motion planning subsystem-local.

\begin{table}[t]
\caption{Comparison of related MAPF and MAPD variants.}
\label{tab:mapd_variant_comp}
\centering
\begingroup
\setlength{\tabcolsep}{2.2pt}
\renewcommand{\arraystretch}{1.1}
\footnotesize
\begin{tabularx}{\columnwidth}{c YYYYY}
\toprule
\textbf{Variants} & \textbf{Online} & \textbf{Multi} & \textbf{Xfer} & \textbf{Buf} & \textbf{TW} \\
\midrule
PDP \cite{coltin_online_2014} &
\cmark{} &
\textcolor{lightgray}{\xmark{}} &
\cmark{} &
\textcolor{lightgray}{\xmark{}} &
\cmark{} \\
\midrule
PERR \cite{ma_multi-agent_2016} &
\textcolor{lightgray}{\xmark{}} &
\textcolor{lightgray}{\xmark{}} &
\cmark{} &
\textcolor{lightgray}{\xmark{}} &
\textcolor{lightgray}{\xmark{}} \\
\midrule
online MAPD \cite{ma_lifelong_2017} &
\cmark{} &
\textcolor{lightgray}{\xmark{}} &
\textcolor{lightgray}{\xmark{}} &
\textcolor{lightgray}{\xmark{}} &
\textcolor{lightgray}{\xmark{}} \\
\midrule
MAPD-D \cite{makino_online_2024} &
\cmark{} &
\textcolor{lightgray}{\xmark{}} &
\textcolor{lightgray}{\xmark{}} &
\textcolor{lightgray}{\xmark{}} &
\cmark{} \\
\midrule
MAPD-MP \cite{flammini_multi-agent_2025} &
\cmark{} &
\cmark{} &
\cmark{} &
\textcolor{lightgray}{\xmark{}} &
\textcolor{lightgray}{\xmark{}} \\
\midrule
MAPD-EA \cite{bonalumi_multi-agent_2025} &
\cmark{} &
\cmark{} &
\textcolor{lightgray}{\xmark{}} &
\textcolor{lightgray}{\xmark{}} &
\textcolor{lightgray}{\xmark{}} \\
\midrule
DPDTW \cite{huang_distributed_2026} &
\cmark{} &
\textcolor{lightgray}{\xmark{}} &
\textcolor{lightgray}{\xmark{}} &
\textcolor{lightgray}{\xmark{}} &
\cmark{} \\
\midrule
MAPF-E \cite{he_conflict-based_2026} &
\textcolor{lightgray}{\xmark{}} &
\cmark{} &
\textcolor{lightgray}{\xmark{}} &
\textcolor{lightgray}{\xmark{}} &
\textcolor{lightgray}{\xmark{}} \\
\midrule
MAPD-HA \cite{flammini_multi_2026} &
\cmark{} &
\cmark{} &
\cmark{} &
\textcolor{lightgray}{\xmark{}} &
\textcolor{lightgray}{\xmark{}} \\
\midrule
MS-MAPD-BHS (ours) &
\cmark{} &
\cmark{} &
\cmark{} &
\cmark{} &
\textcolor{lightgray}{\xmark{}} \\
\bottomrule
\end{tabularx}
\par\smallskip
\footnotesize\textbf{Online}: Online tasks.
\textbf{Multi}: multiple fleets/environments.
\textbf{Xfer}: handover/exchange.
\textbf{Buf}: buffer/storage.
\textbf{TW}: time windows/deadlines.
\endgroup
\end{table}

\section{Problem Statement}
\label{sec:problem}

We formally define \emph{MS-MAPD-BHS}: \textbf{M}ulti-\textbf{S}ubsystem MAPD with \textbf{B}uffer-limited \textbf{H}andover \textbf{S}tations. We study the two-subsystem, one-way-flow case, which is the smallest setting in which a shared finite buffer induces blocking and starvation. The controller and experiments are limited to this case; reverse flow, bidirectional competition for the same station, and chains with more than two subsystems are outside the evaluated scope. Throughout this section, we use the notation summarized in Table \ref{tab:notation}.

\begin{table}[b]
\caption{Selected notation used in Sec. \ref{sec:problem}.}
\label{tab:notation}
\centering
\small
\begin{tabular}{ll}
\toprule
Symbol & Meaning \\
\midrule
$G=(V,E)$ & global undirected graph environment \\
$V_i\subseteq V$ & vertex set of subsystem $i$ \\
$G_i=(V_i,E_i)$ & subsystem-$i$ induced subgraph $G[V_i]$ \\
$H\subseteq V$ & handover station vertices \\
$A=\bigcup_i A_i$ & set of all agents (for all subsystems $i$) \\
$p_a(t)$ & vertex of agent $a$ at timestep $t$ \\
$\tau^{\st{load}},\tau^{\st{unload}}$ & load/unload durations [timesteps] \\
$B_h$ & buffer capacity of dock $h\in H$ \\
$b_h(t)$ & buffer occupancy of dock $h$ at $t$ \\
$o=(s(o),g(o),r(o))$ & task (source, destination, release time) \\
$c(o)$ & completion time of task $o$ \\
\bottomrule
\end{tabular}
\end{table}

\subsection{Environment and motion model}
As in the standard discrete MAPF/MAPD setting, the environment is an undirected graph
$G=(V,E)$ typically induced by a 4-neighbor grid map: each free grid cell is a vertex and an edge connects
horizontally/vertically adjacent free cells. Time is discrete, i.e., $t\in\bbZ_{\ge 0}$.

Each subsystem $i \in \{1, 2\}$ operates within its own dedicated region defined by a vertex set $V_i \subseteq V$, forming a vertex-induced subgraph $G_i = G[V_i] = (V_i, E_i)$, with $E_i=\left\{\{u, v\} \in E: u, v \in V_i\right\}$. These subsystems are geographically disjoint except for a set of \emph{shared dock vertices}, $H=\{h_1,\dots,h_{|H|}\}$, which constitute their only overlap ($V_1\cap V_{2} = H$) and are otherwise disjoint ($(V_1\setminus H)\cap (V_2\setminus H)=\emptyset$). Crucially, there are no edges connecting non-dock regions directly across subsystems. This means payload transfers are exclusively conducted via dock vertices in $H$. Each dock vertex itself represents an isolated station, implying no direct edges exist between distinct dock vertices ($h \neq h^{\prime} \in H \implies \{h, h^{\prime}\} \notin E$).

Each subsystem $i$ has its own set of agents $A_i$. Let $p_a(t)$ denote the location (i.e., occupied vertex) of agent $a$ at the timestep $t$. Agents are strictly confined to their respective $G_i$: for any $a\in A_i$, $p_a(t) \in V_i, \, \forall t$, and all moves must be along edges in $E_i$.
Also, at each $t$, an agent executes one action (either wait or move to an adjacent vertex in its subsystem graph), yielding the next position $p_a(t+1)\in \{p_a(t)\}\cup N_{G_i}(p_a(t))$, where $N_{G_i}(.)$ denotes neighbor set in $G_i$. All agents act simultaneously at each $t$. Let $A=\bigcup_i A_i$. A \emph{vertex collision} between agents $a\neq b\in A$ occurs at timestep $t$ iff
$p_a(t)=p_b(t)$. An \emph{edge collision} (swap) occurs at timestep $t$ iff
$p_a(t)=p_b(t+1)$ and $p_a(t+1)=p_b(t)$. Agents must avoid collisions. We assume perfect execution.

We define $\tau^{\st{load}},\tau^{\st{unload}}\in\bbZ_{>0}$ as the fixed durations for load/unload operations at task sources/destinations or dock vertices. Operations are non-preemptive and are modeled as consecutive waits at the operation vertex. If an agent arrives at time $t_0$ and waits for $\tau$ steps, it occupies that vertex at integer times $\{t_0, \ldots, t_0+\tau\}$ and may depart at $t_0+\tau+1$. An agent begins its assigned operation immediately upon arrival at the vertex if preconditions are met (see Section~\ref{subsec_buffer}).

\subsection{Handover station interface model}
Subsystems exchange payloads only via handover stations. Each station is represented by a shared
\emph{dock vertex} $h\in H$ together with a finite buffer of capacity $B_h\in\bbZ_{>0}$ (max number of storable payloads). At any timestep $t$, the dock vertex $h$ may be occupied by at most one agent.

\subsection{Tasks and buffer dynamics}
\label{subsec_buffer}
Tasks arrive online. Each task $o$ is a triple $(s(o),g(o),r(o))$ with source $s(o)\in V_1\setminus H$, destination $g(o)\in V_2\setminus H$, and release time $r(o)\in\bbZ_{\ge 0}$. Thus subsystem~1 is \emph{upstream} and subsystem~2 is \emph{downstream}. We consider only one-way inter-subsystem tasks from upstream to downstream; reverse and bidirectional flows are not modeled in this paper. At time $r(o)$, a payload becomes available at $s(o)$ and remains there until an upstream agent loads it. Payloads do not block agent paths.

The buffer models storage capacity only, holding up to $B_h$ payloads. By imposing no physical ordering (e.g., FIFO/LIFO), it abstracts a common staging area where any item can be accessed. Buffered payloads retain their task association $o$ (including destination $g(o)$). A downstream agent loading at $h$ may retrieve any available payload, providing maximum planning flexibility but complicating coordination. Each agent starts empty and carries at most one payload at a time.

Let $b_h(t)\in\mathbb{Z}_{\geq0}$ denote the buffer occupancy at dock vertex $h$ at timestep $t$. All buffers are initially empty (i.e., $b_h(0) = 0, \forall h \in H$), and occupancy is always limited by the buffer capacity (i.e., $0 \leq b_h(t) \leq B_h$ for all $t$). An upstream agent performs the following sequence: loads a payload at $s(o)$ (duration $\tau^{\st{load}}$), travels to a chosen dock vertex $h\in H$, and unloads at $h$ (duration $\tau^{\st{unload}}$). An upstream agent may enter $h$ and start the unload operation at timestep $t$ only if the buffer is not full ($b_h(t) < B_h$). If the operation starts at $t$, the payload enters the buffer, and $b_h$ increases by one, at timestep $t + \tau^{\st{unload}}$.

A downstream agent performs the following sequence: loads a buffered payload at $h$ (duration $\tau^{\st{load}}$), travels to the corresponding $g(o)$, and unloads it (duration $\tau^{\st{unload}}$). A downstream agent may enter $h$ and start the load operation at timestep $t$ only if the selected payload is present at $h$; in particular, $b_h(t) > 0$. If the operation starts at $t$, that payload is removed from the buffer, and $b_h$ decreases by one at timestep $t + \tau^{\st{load}}$. The entry guards for upstream and downstream operations apply only at the planned dock-entry time: an agent may wait elsewhere or replan until the operation becomes admissible.

\subsection{Service time metric and optimization goal}
Let $c(o)$ denote the completion time of task $o$, defined as the timestep at which the downstream agent finishes its unload operation at $g(o)$. For a horizon $T$, let $O_T=\{o:c(o)\leq T\}$. When $O_T\neq\emptyset$, service time \cite{flammini_multi-agent_2025} is  the mean completion delay: 
\begin{equation}
\mathrm{ServiceTime}(T)
=
\frac{1}{\left|O_T\right|}
\sum_{o \in O_T} \bigl( c(o) - r(o) \bigr),
\label{eq:service_time}
\end{equation}

Given $\left(G,\left\{A_i\right\},\left\{B_h\right\}, \tau^{\st{load}}, \tau^{\st{unload}} \right)$ and an online task stream, a solution to MS-MAPD-BHS is an online policy that assigns tasks, selects handover stations, schedules station usage, and plans collision-free paths. The primary objective is low service time subject to motion and buffer safety; Sec.~\ref{sec:experiments} additionally reports throughput and backlog to diagnose overload.

\section{Handover-Aware Reservation and Routing (HARR)}
\label{sec:controller}

\subsection{Overview and Token Passing}

Token Passing (TP) is an online MAPD scheme in which a token stores every local agent's committed, time-indexed path and task assignment~\cite{ma_lifelong_2017}. An agent requests the token when it reaches the end of its stored path. While holding the token, it selects an unassigned task, computes a space-time path that avoids the other paths stored in the token, writes its new path back, and releases the token. We call this stored sequence a \emph{token path}; for agent $a$, it is $\pi_a=(\pi_a(t),\pi_a(t{+}1),\ldots)$ with $\pi_a(t)=p_a(t)$.

HARR maintains one TP token per subsystem and a shared station manager that reserves dock occupancy and enforces buffer feasibility. Each inter-subsystem task is decomposed into two subtasks, called \emph{legs}: an upstream \emph{push leg} from $s(o)$ to a selected dock $h\in H$, followed by a downstream \emph{pull leg} from the same dock to $g(o)$.
HARR processes pulls before pushes at each timestep to favor buffer drainage. Alg.~\ref{alg:harr} details the HARR online control loop.

Throughout this section, $\dist_{G_i}(u,v)$ denotes the shortest-path distance between $u,v\in V_i$ in $G_i$.

\begin{figure}[t]
  \centering
  \includegraphics[width=\columnwidth]{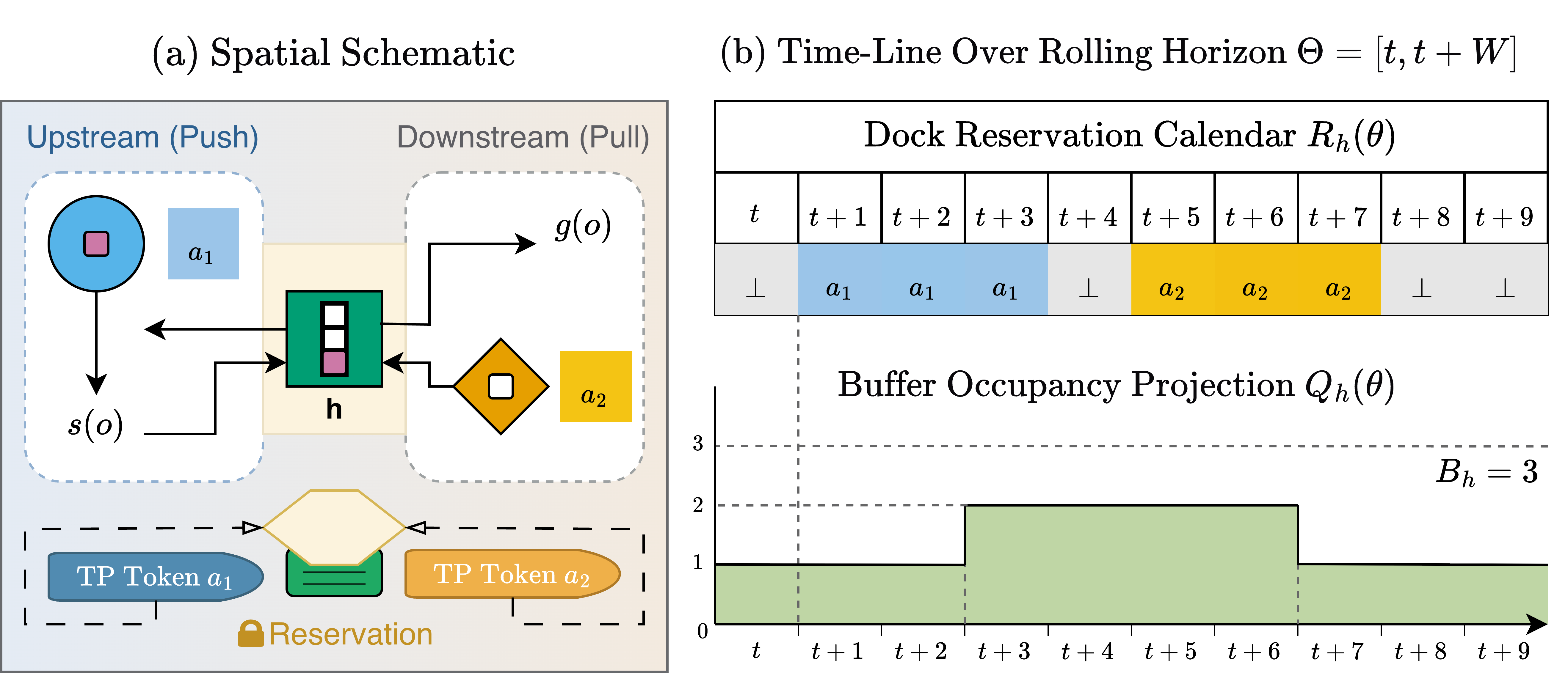}
  \caption{HARR runs TP in each subsystem and couples the planners only through shared station state. $R_h$ reserves exclusive dock occupancy, and $Q_h$ is a deterministic projection obtained by applying the events of already committed buffer operations to the observed current occupancy.}
  \label{fig:harr_overview}
\end{figure}

\subsection{Shared station calendars}
At global time $t$, HARR maintains per-dock calendars over a rolling horizon
$\Theta=[t,t{+}W]\cap\mathbb{Z}$, where $W\in\bbZ_{>0}$ is the window size.

\paragraph{Dock reservation calendar}
For each dock vertex $h\in H$, we define its dock reservation calendar as a time-indexed map $R_h: \bbZ_{\ge 0} \to A\cup\{\bot\}$. $R_h(\theta)\in A$ indicates the unique agent reserved to occupy $h$ at time $\theta$, or $R_h(\theta)=\bot$ if $h$ is unreserved.
This calendar $R_h$ is global; it is queried during planning and updated when a new token path is committed
(Alg.~\ref{alg:harr}, \textsc{ReserveDocks}).

\paragraph{Committed buffer-event calendar and occupancy projection}
For each dock $h$, the event calendar $E_h: \bbZ_{\ge 0} \to \mathbb{Z}$ stores occupancy changes caused by already committed operations. An upstream unload beginning at dock-entry time $t_0$ adds $+1$ to $E_h(t_0+\tau^{\st{unload}})$, whereas a downstream load adds $-1$ to $E_h(t_0+\tau^{\st{load}})$. These events define the deterministic occupancy projection $Q_h$:
\begin{equation}
Q_h(t)=b_h(t),\,\, Q_h(\theta)=Q_h(\theta-1)+E_h(\theta)\;\;\text{for }\theta>t.
\label{eq:projection}
\end{equation}
Thus, $Q_h$ is the time-indexed buffer schedule implied by accepted operations. As the planning horizon advances, past events are discarded and $Q_h(t)$ is re-anchored to the observed occupancy $b_h(t)$. Events are committed online via \textsc{CommitDockAndBuffer} (Alg.~\ref{alg:harr}).

\paragraph{Dock occupancy interval}
Under the operation model defined in Sec.~\ref{sec:problem}, an operation of duration $\tau$ that begins upon an agent entering $h$ at time $t_0$ \emph{exclusively}
occupies the dock vertex at integer times
\[
I^{\st{dock}}(t_0,\tau)=\{t_0,t_0{+}1,\dots,t_0{+}\tau\}.
\]

\paragraph{Station-feasible insertion of a dock operation}
To insert an operation for agent $a$ at dock $h$ with dock-entry time $t_0$, duration $\tau$, and event sign
$\sigma\in\{+1,-1\}$ ($+1$ for push/unload, $-1$ for pull/load), we define the event time $\theta^{\st{evt}}:=t_0+\tau$ with $\theta^{\st{evt}}\in\Theta$.
The insertion is station-feasible iff:
\begin{align}
&\forall \theta\in I^{\st{dock}}(t_0,\tau):\;\;R_h(\theta)=\bot, \label{eq:dock_free}\\
&\text{start guard: }\;\;
\begin{cases}
Q_h(t_0)<B_h & \text{if }\sigma=+1\ (\text{push})\\
Q_h(t_0)>0   & \text{if }\sigma=-1\ (\text{pull})
\end{cases} \label{eq:start_guard}\\
&\forall \theta\in[\theta^{\st{evt}},t{+}W]\cap\mathbb{Z}:\;\;
0\le Q_h(\theta)+\sigma \le B_h. \label{eq:cap_guard}
\end{align}
\noindent These conditions require:
\begin{itemize}
    \item Condition~\eqref{eq:dock_free}: The dock must be unreserved ($R_h(\theta)=\bot$) throughout the operation's exclusive interval $I^{\st{dock}}(t_0,\tau)$.
    \item Condition~\eqref{eq:start_guard}: At $t_0$, the buffer must have space for a push ($Q_h(t_0)<B_h$) or contain the selected pull payload; the latter also implies $Q_h(t_0)>0$.
    \item Condition~\eqref{eq:cap_guard}: After inserting the candidate event, the committed occupancy projection must remain within $[0, B_h]$ through the end of the enforced horizon, so the insertion cannot invalidate an already accepted later operation.
\end{itemize}

\subsection{Station-aware Token Passing}
When an agent requests its subsystem token, \textsc{PlanFirstFeasible} runs space-time A* on $G_i$. The search avoids vertex and edge conflicts with the other token paths in that subsystem, and additionally respects the shared station calendar.

\paragraph{Dock constraints}
During A* for agent $a$, a dock state $(h,\theta)$ is unavailable whenever $R_h(\theta)\notin\{\bot,a\}$.
After a path is accepted, \textsc{ReserveDocks} writes every dock occupancy on that path within the horizon to $R_h$, preventing cross-subsystem conflicts at shared dock vertices.

\paragraph{Dock operations}
A dock operation beginning $t_0$ is admissible only if Eqs.~\eqref{eq:dock_free}--\eqref{eq:cap_guard} hold. If accepted, the corresponding event is posted to $E_h$ and the occupancy projection $Q_h$ is updated before the next request is considered.

\subsection{Leg paths and unassigned-leg sets}
\label{subsec:leg_templates}
Let $\mathcal{E}_i\subseteq V_i$ denote the endpoint set used by TP in subsystem $i$
(task endpoints plus additional parking endpoints). Parking is restricted to $\mathcal{E}_i\setminus H$.

HARR uses two sets to track work that is currently available for assignment.
$T^{\uparrow}(t)$ contains released tasks whose push leg is unassigned, and
$T^{\downarrow}(t)$ contains buffered tasks whose pull leg is unassigned.
For each $o\in T^{\downarrow}(t)$, the data structure also stores the dock $h(o)$ where the payload is present. A task leaves $T^{\uparrow}$ when its push path is committed, enters $T^{\downarrow}$ only after the upstream unload completes, leaves $T^{\downarrow}$ when its pull path is committed, and exits the system after the downstream unload at $g(o)$. While a leg is in progress, its state is represented by the responsible agent's token path rather than duplicated in either set.

\paragraph{Push template (subsystem 1)}
A push for task $o$ to dock $h$ plans through
$s(o)$ (wait $\tau^{\st{load}}$), then $h$ (wait $\tau^{\st{unload}}$), and finally a parking endpoint $e\in\mathcal{E}_1\setminus H$.
The dock operation uses $(\sigma,\tau)=(+1,\tau^{\st{unload}})$.

\paragraph{Pull template (subsystem 2)}
A pull for buffered task $o$ at dock $h(o)$ plans through
$h(o)$ (wait $\tau^{\st{load}}$), then $g(o)$ (wait $\tau^{\st{unload}}$), and finally a parking endpoint $e\in\mathcal{E}_2\setminus H$.
The dock operation uses $(\sigma,\tau)=(-1,\tau^{\st{load}})$.

\begin{algorithm}[t]
\caption{HARR: station- and buffer-aware decoupled Token Passing.}
\label{alg:harr}
\small
\begin{algorithmic}[1]
\Require Subsystem graphs $G_i$, agents $A_i$, docks $H$, capacities $B_h$, endpoints $\mathcal{E}_i$
\Require Operation times $\tau^{\st{load}},\tau^{\st{unload}}$, parameters $W$, $k_{\max}$, $M_{\max}$
\State Initialize TP tokens: $\pi_a \gets [p_a(0)]$ for all $a\in A$
\State Initialize station calendars $R_h(\theta)\leftarrow\bot$, $E_h(\theta)\leftarrow 0$ for all $h\in H$, $\theta\in[0,W]$
\State Initialize unassigned-leg sets $T^{\uparrow}\gets\emptyset$, $T^{\downarrow}\gets\emptyset$
\For{$t=0,1,2,\dots$}
    \State Observe $b_h(t)$ for all $h\in H$
    \State Shift horizon $\Theta=[t,t{+}W]\cap\bbZ$; discard $R_h(\theta),E_h(\theta)$ for $\theta<t$
    \State Anchor $Q_h(t)=b_h(t)$ and recompute $Q_h$ from $E_h$ on $\Theta$
    \State Insert newly released tasks into $T^{\uparrow}$
    \State Update the sets from executed operations (completed push $\rightarrow T^{\downarrow}$ with $h(o)$; completed pull $\rightarrow$ remove from system)
    \State \textsc{ProcessRequests}$(2,\textsc{Pull})$ \Comment{downstream first}
    \State \textsc{ProcessRequests}$(1,\textsc{Push})$
    \State Execute one timestep: all agents follow their token paths; advance to $t{+}1$
\EndFor

\Function{\textsc{ProcessRequests}}{$i,\textsc{Mode}$}
    \State $(T^\star,\sigma,\tau)\gets \textsc{ModeParams}(\textsc{Mode})$ \Comment{$T^\star$: unassigned legs for this mode}
    \For{each agent $a\in A_i$ requesting the token}
        \State $\mathcal{C}\gets \textsc{Candidates}(a,\textsc{Mode},T^\star)$
        \State $(\pi,o,h,t_0)\gets \textsc{PlanFirstFeasible}(a,\mathcal{C},\textsc{Mode},R,Q)$
        \If{$\pi\neq\bot$}
            \State $\pi_a\gets\pi$; \textsc{ReserveDocks}$(a,\pi,R)$
            \State \textsc{CommitDockAndBuffer}$(h,t_0,\sigma,\tau,E,Q)$; remove $o$ from $T^\star$ \Comment{the leg is now in progress}
        \Else
            \State $\pi_a\gets \textsc{Park}(a,\mathcal{E}_i\setminus H)$
        \EndIf
    \EndFor
\EndFunction
\end{algorithmic}
\end{algorithm}

\paragraph{Algorithm helpers}
\textsc{ModeParams} selects $(T^{\uparrow},+1,\tau^{\st{unload}})$ for pushes and $(T^{\downarrow},-1,\tau^{\st{load}})$ for pulls; \textsc{Candidates} applies the ordering below. \textsc{PlanFirstFeasible} runs space-time A* for each pair and enforces $R$ and Eqs.~\eqref{eq:dock_free}--\eqref{eq:cap_guard}. \textsc{ReserveDocks} writes accepted dock-time states to $R$, and \textsc{CommitDockAndBuffer} posts the event to $E$ and updates $Q$. \textsc{Park} uses the same path constraints to reach a non-dock endpoint.

\subsection{Task and dock selection}
When a downstream agent $a\in A_2$ requests the token, it considers up to $M_{\max}$ candidates
$o\in T^{\downarrow}$ in increasing order of
\[
J_{\downarrow}(a,o)=\dist_{G_2}\bigl(p_a(t),h(o)\bigr)+\dist_{G_2}\bigl(h(o),g(o)\bigr).
\]
When an upstream agent $a\in A_1$ requests the token, it considers up to $M_{\max}$ tasks $o\in T^{\uparrow}$ (oldest first).
For each task, it considers up to $k_{\max}$ docks $h\in H$ in increasing order of
\[
J_{\uparrow}(a,o,h)=
\dist_{G_1}\bigl(p_a(t),s(o)\bigr)
+\dist_{G_1}\bigl(s(o),h\bigr)
+\dist_{G_2}\bigl(h,g(o)\bigr).
\]
Candidates are attempted in the above order; the first candidate for which station-aware A* finds a feasible plan is accepted.

\subsection{Computational cost, guarantees, and feasibility}

\paragraph{Computational cost}
For one token request, a pull attempts at most $M_{\max}$ task--dock pairs, whereas a push attempts at most $M_{\max}\min\{k_{\max},|H|\}$ pairs. Each pair invokes one space-time A* search. In addition to TP's collision checks, HARR examines the $\tau+1$ dock slots in Eq.~\eqref{eq:dock_free} and at most $W+1$ occupancy entries in Eq.~\eqref{eq:cap_guard}. The rolling-horizon arrays $R$, $E$, and $Q$ each store $|H|(W+1)$ entries. Enumeration stops at the first feasible pair.

\paragraph{Collision safety}
Within each subsystem, standard TP checks preserve vertex- and edge- collision freedom among committed local paths. Across subsystems, the only shared vertices are docks, and $R_h(\theta)$ permits at most one agent to occupy a dock at a time. Therefore, under perfect execution, committed dock use is collision-free over the enforced reservation horizon. It does not claim completeness or collision safety beyond $t+W$, where future paths have not yet been committed.

\paragraph{Sequential insertion semantics}
Requests are processed sequentially as in Alg.~\ref{alg:harr}.
After each accepted plan, \textsc{CommitDockAndBuffer} immediately updates $(E_h,Q_h)$
before the next candidate is checked. The guarantee below therefore assumes this insertion order and perfect execution.

\paragraph{Buffer safety}
At time $t$, each occupancy projection is anchored at $Q_h(t)=b_h(t)\in[0,B_h]$. Committing an operation leaves $Q_h$ unchanged before its event and shifts it by $\sigma$ from the event time onward. Condition~\eqref{eq:cap_guard} is exactly the requirement that the updated projection remain in $[0,B_h]$. By induction over sequentially accepted insertions,
every committed event calendar is occupancy-consistent over $[t,t+W]$. The guarantee is horizon-local. HARR makes no claim beyond $t+W$ until the horizon advances and the projection is re-anchored.

\paragraph{Feasibility and completeness}
MS-MAPD-BHS is not universally solvable. Feasibility depends on subsystem
geometry, endpoint availability, reachability between task vertices and admissible docks, staffing balance, and interface load. HARR is not complete: it uses a finite horizon, caps task candidates at $M_{\max}$, and inherits the limitations of prioritized TP planning. Our domains avoid trivial endpoint deadlocks---agents can park outside $H$, and the relevant task and dock vertices are reachable---but a full completeness and stability characterization is left to future work.
\section{Comparators and Ablations}
\label{sec:baselines}

We compare \textsc{HARR} with one coupled reference method and two mechanism ablations. The reference evaluates the computational effect of subsystem-local rather than global Token Passing, whereas the ablations isolate online dock selection and forward buffer projection.

\paragraph{\textsc{MonoSBTP} (coupled station- and buffer-aware TP)}
We run TP with a single global token over all agents $A$, while restricting each agent's search space to its subsystem vertices $V_i$. Dock exclusivity is enforced implicitly by global collision avoidance at shared dock vertices, and buffer feasibility uses the same committed occupancy projection $(E_h,Q_h)$ and capacity guard (Eq.~\eqref{eq:cap_guard}) as \textsc{HARR}.

\paragraph{\textsc{FixedDock} (no online dock selection)}
This method is identical to \textsc{HARR}, except that each task is assigned at release to the dock minimizing $\dist_{G_1}(s(o),h)+\dist_{G_2}(h,g(o))$ and cannot change docks online. It is an ablation of adaptive dock selection.

\paragraph{\textsc{StartResBuf} negative control (local start-time admission only)}
This ablation omits the rolling-horizon occupancy projection: a newly inserted dock operation is checked only against dock availability and its local start guard, without $(E_h,Q_h)$ and Eq.~\eqref{eq:cap_guard}. Because online insertions can occur out of chronological order, a newly accepted earlier event can invalidate an already committed later operation. In our simulator, \textsc{StartResBuf} yielded no valid run in the tested settings, so it is retained as a negative control and excluded from quantitative plots.

\begin{table}[t]
\caption{Comparators and ablations.}
\label{tab:baseline_summary_main}
\centering
\small
\setlength{\tabcolsep}{5pt}
\begin{tabular}{lcccc}
\toprule
Method & G-TP & $R$ & Projection & DynDock \\
\midrule
\textsc{MonoSBTP}    & \cmark{} & \textcolor{lightgray}{\xmark{}} & \cmark{} & \cmark{} \\
\textsc{FixedDock}   & \textcolor{lightgray}{\xmark{}} & \cmark{} & \cmark{} & \textcolor{lightgray}{\xmark{}} \\
\textsc{StartResBuf} & \textcolor{lightgray}{\xmark{}} & \cmark{} & \textcolor{lightgray}{\xmark{}} & \cmark{} \\
\midrule
\textsc{HARR} (ours) & \textcolor{lightgray}{\xmark{}} & \cmark{} & \cmark{} & \cmark{} \\
\bottomrule
\end{tabular}
\par\smallskip
\footnotesize
\textbf{G-TP}: token scope in \textsc{ProcessRequests}. \\
\textbf{$R$}: written by \textsc{ReserveDocks}. \\
\textbf{Projection}: updated by \textsc{CommitDockAndBuffer}. \\
\textbf{DynDock}: dock enumeration in \textsc{Candidates}.
\end{table}

\section{Experiments}
\label{sec:experiments}

We evaluate \textsc{HARR} in a C++ simulator implementing the two-subsystem \emph{MS-MAPD-BHS} model from Sec.~\ref{sec:problem}. Unless stated otherwise, all methods use the same motion model, task process, and endpoint sets $\mathcal{E}_i$ (with parking restricted to $\mathcal{E}_i\setminus H$), as well as uniform station parameters $\tau^{\st{load}}=\tau^{\st{unload}}=2$ and $B_h=B$. For \textsc{HARR}, we set $W=60$, $M_{\max}=20$, and $k_{\max}=8$. Because the largest tested configuration has $|H|=8$, the dock-candidate cap never binds in the reported experiments. Default provisioning is $|H|=4$, $B=1$, with $n_1=n_2=12$ agents.

\subsection{Domains and station layouts}
\label{subsec:exp_domains}

We use two representative grid domains:
\texttt{Empty}, an open map ($40\times20$) that isolates interface-induced effects (dock contention, blocking, and starvation); and
\texttt{Warehouse}, a shelf-like layout ($34\times26$) that induces within-subsystem congestion and longer detours.

At each timestep $t$, the number of new inter-subsystem tasks is $N_t\sim \mathrm{Poisson}(\rho)$.
Each task $o=(s(o),g(o),r(o))$ samples $s(o)$ uniformly from a fixed upstream source set
$\mathcal{S}\subset V_1\setminus H$ and $g(o)$ uniformly from a fixed downstream destination set
$\mathcal{G}\subset V_2\setminus H$. We choose 4 sources and 4 destinations, whose locations are initialized by random seeds.

Handover stations are distributed approximately uniformly along the interface; the released configurations specify their exact coordinates for every tested $|H|$.  To compare runs across different numbers of docks $|H|$ and operation times, we define the normalized offered dock load
\[
\varrho \;=\; \rho \cdot \frac{(\tau^{\st{unload}}{+}1)+(\tau^{\st{load}}{+}1)}{|H|},
\]
where $\tau{+}1$ matches the dock-occupancy interval length $|I^{\st{dock}}(\cdot,\tau)|$ used by \textsc{HARR}
(Sec.~\ref{sec:controller}). Thus, $\varrho$ compares requested dock time with nominal aggregate dock capacity. It is an offered-load indicator, not a sufficient stability condition, because path congestion, staffing imbalance, and heterogeneous dock accessibility also affect throughput.

\subsection{Protocol and metrics}
\label{subsec:exp_protocol}

Each configuration runs for $T_{\st{tot}}=4000$ timesteps. We discard the first $T_{\st{warm}}=1000$ timesteps (25\% of the run) to reduce the common initialization transient caused by initially empty buffers and seeded initial agent positions; the same warm-up is used for every method and configuration.
Results are averaged over $N_{\st{seed}}=10$ independent random seeds and reported with 95\% confidence intervals.

\paragraph{Performance metrics}
We report three complementary metrics:
\begin{enumerate}
  \item Service time [timesteps]. $\mathrm{ServiceTime}(T_{\st{tot}})$ from Eq.~\eqref{eq:service_time}, computed over tasks released
  after warm-up and completed by $T_{\st{tot}}$.
  \item Throughput [tasks/timestep] (stability).
  \[
  \mathrm{Throughput} \;=\; \frac{\left|\left\{o:\; c(o)\in (T_{\st{warm}},T_{\st{tot}}]\right\}\right|}{T_{\st{tot}}-T_{\st{warm}}},
  \]
  i.e., completed tasks per timestep during the measurement window.
  \item Backlog [tasks] (unserved demand).
\[
\mathrm{Backlog}
=
\left|\left\{o:\; T_{\st{warm}}<r(o)\le T_{\st{tot}},\; c(o)>T_{\st{tot}}\right\}\right|.
\]
\end{enumerate}
Throughput and backlog remain informative in overload regimes and prevent survivorship bias.

\subsection{Results}
\label{subsec:exp_results}

The two comparison studies target different mechanisms. The load and provisioning sweeps compare \textsc{HARR} with the \textsc{FixedDock} ablation to isolate adaptive dock selection, whereas the fleet-size sweep compares \textsc{HARR} with the coupled \textsc{MonoSBTP} reference to isolate global-token coupling. They should therefore be read as two controlled comparisons, not as a single three-way ranking.

\subsubsection{Adaptive dock selection: \textsc{HARR} vs.\ \textsc{FixedDock}}
\label{subsec:exp_results_perf}

We first study whether online dock selection improves both service time and stability as the interface approaches saturation. We sweep the offered dock load $\varrho$ on both maps under default provisioning and balanced staffing.
For each configuration, we report service time, throughput, and backlog.

\begin{figure*}[t]
\centering
\includegraphics[width=\textwidth]{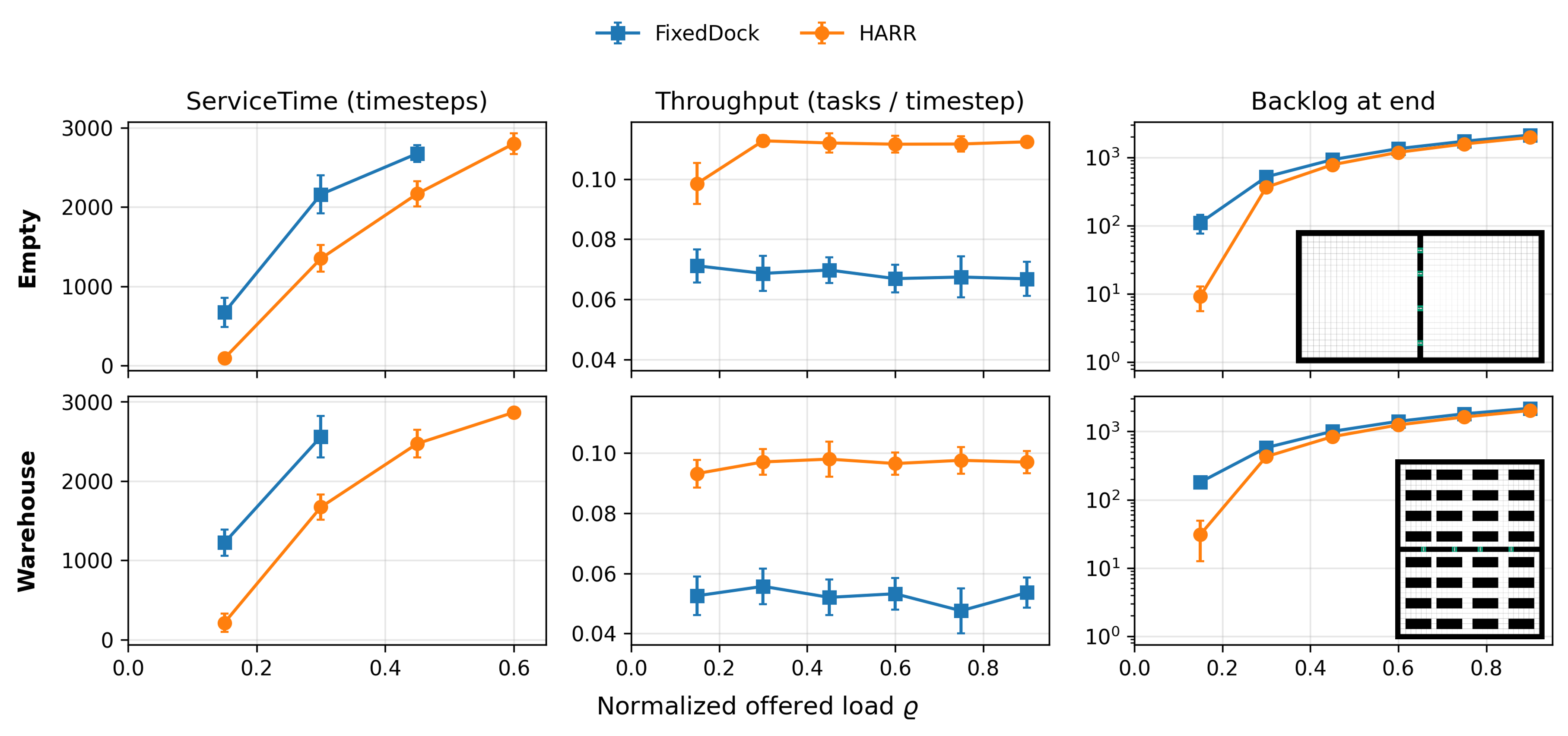}
\caption{Load sweep under default provisioning ($|H|=4$, $B=1$). Rows are \texttt{Empty} and \texttt{Warehouse}; columns show service time [timesteps] (lower is better), throughput [tasks/timestep] (higher is better), and backlog [tasks] (lower is better, log scale).}
\label{fig:load_sweep_metrics}
\end{figure*}

Figure~\ref{fig:load_sweep_metrics} shows that \textsc{HARR} outperforms \textsc{FixedDock} across the tested load range. At moderate load, it achieves higher throughput and substantially lower backlog on both maps (Table~\ref{tab:key_load}). Near saturation, throughput and backlog reveal overload more reliably than service time; the largest gains occur on \texttt{Warehouse}, where congestion makes static dock assignments less effective.

\begin{table*}[t]
\caption{Moderate-load results ($\varrho=0.15$; default provisioning). Throughput is measured in tasks/timestep and backlog in tasks.}
\label{tab:key_load}
\centering
\small
\setlength{\tabcolsep}{4pt}
\begin{tabular}{lccc|ccc}
\toprule
& \multicolumn{3}{c}{Throughput $\uparrow$} & \multicolumn{3}{c}{Backlog $\downarrow$} \\
Map & \textsc{FixedDock} & \textsc{HARR} & Gain & \textsc{FixedDock} & \textsc{HARR} & Reduction \\
\midrule
Empty &
0.071$\pm$0.005 & $\mathbf{0.099\pm0.007}$ & +39\% &
110$\pm$34 & $\mathbf{9.2\pm3.6}$ & 92\% \\
Warehouse &
0.053$\pm$0.006 & $\mathbf{0.093\pm0.005}$ & +77\% &
181$\pm$33 & $\mathbf{31\pm18}$ & 83\% \\
\bottomrule
\end{tabular}
\end{table*}

\begin{figure}[t]
\centering
\includegraphics[width=\columnwidth]{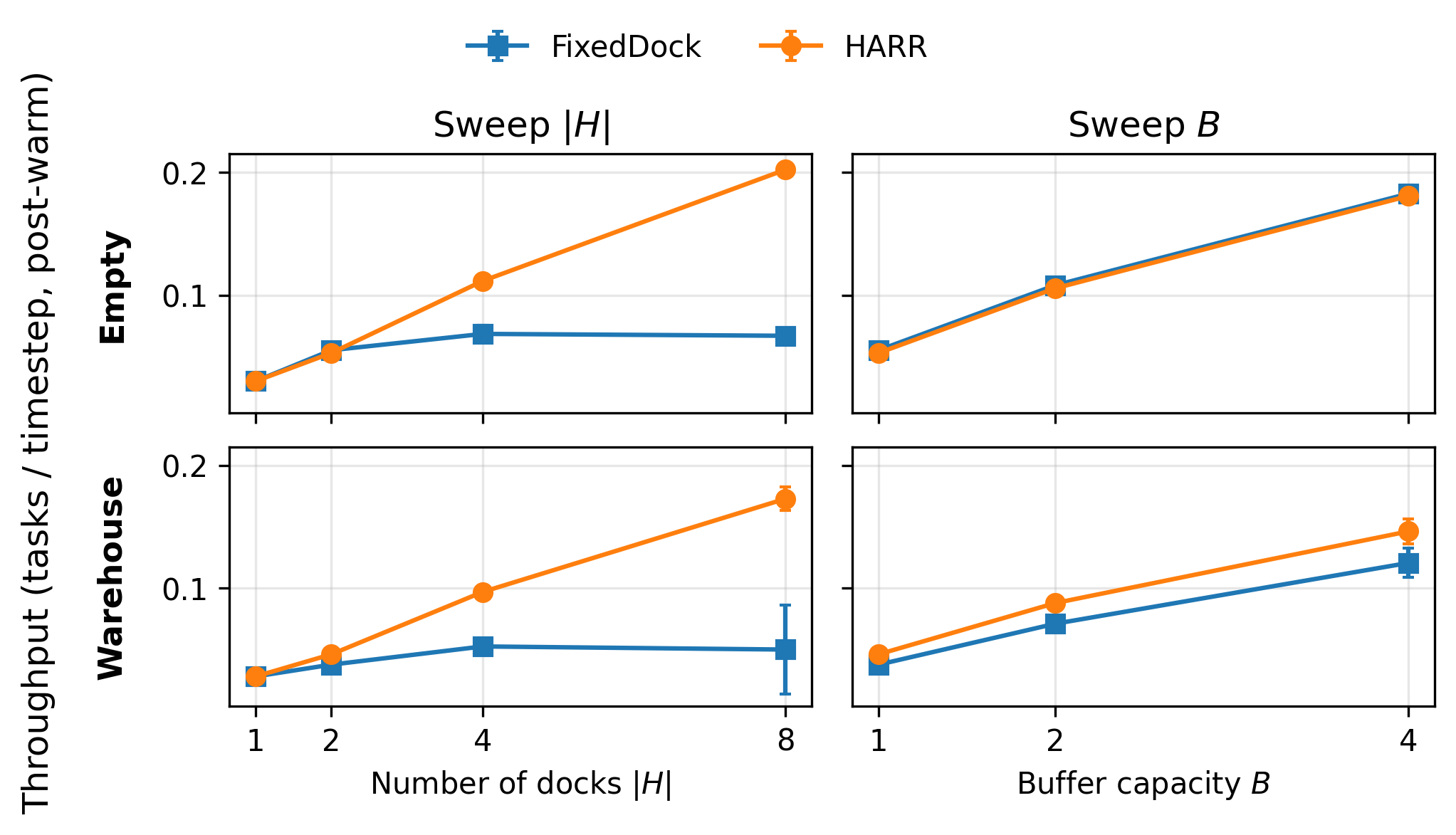}
\caption{Interface provisioning sweeps (throughput). Columns sweep the number of docks $|H|$ and the buffer capacity $B$.}
\label{fig:provisioning_throughput}
\end{figure}

We next vary the number of docks $|H|$ and buffer capacity $B$ separately at an arrival rate near saturation. Figure~\ref{fig:provisioning_throughput} shows that \textsc{HARR} benefits more from additional docks because it can exploit the currently reachable, buffer-feasible interface. At $|H|=8$, its throughput is 202\% higher on \texttt{Empty} and 245\% higher on \texttt{Warehouse}.

\subsubsection{Fleet-size planning cost: \textsc{HARR} vs.\ \textsc{MonoSBTP}}
\label{subsec:exp_results_scale}

We compare the computational cost of decoupled planning in \textsc{HARR} with the coupled  \textsc{MonoSBTP} reference while checking service time. We sweep the total number of agents $N\in \{5,10,15,20,25,30,35,40,45\}$ with a balanced split $(n_1,n_2)=(\lfloor N/2\rfloor,\lceil N/2\rceil)$. Workload, workspace, and interface provisioning remain fixed, so this is a \emph{fleet-size} study rather than a workspace-size scaling study.

\begin{figure}[t]
\centering
\includegraphics[width=\columnwidth]{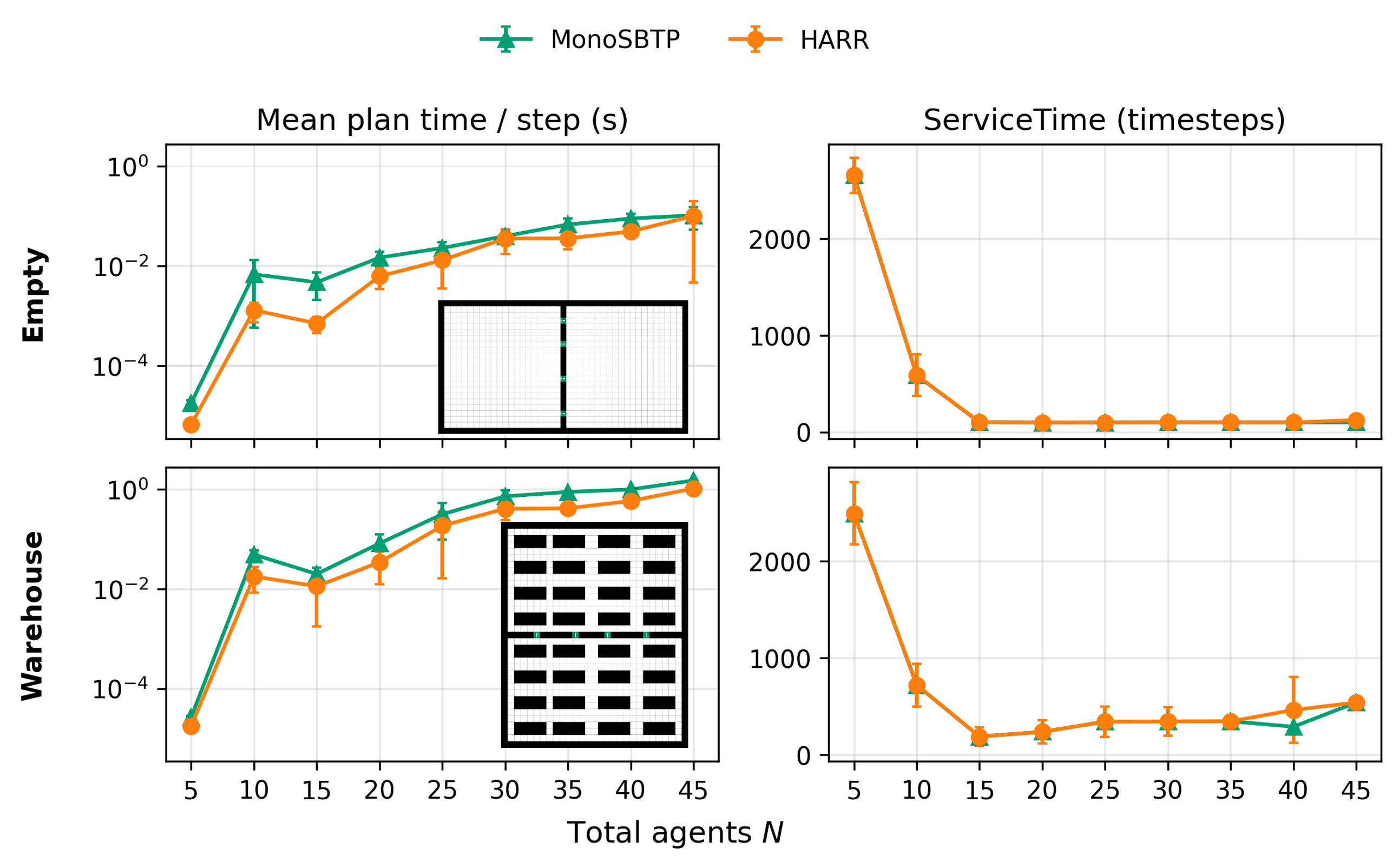}
\caption{Fleet-size sweep on two fixed workspaces. Mean planning time per step [s] and service time [timesteps] are both lower when better; curves summarize ten seeds.}
\label{fig:scalability_overview}
\end{figure}

Figure~\ref{fig:scalability_overview} shows that \textsc{HARR} has lower or comparable mean planning time across the tested fleet sizes, with the clearest advantage on the structured \texttt{Warehouse} map. Service time is non-monotonic: adding agents initially reduces delay, but larger fleets eventually create more path interference and coordination work. On \texttt{Warehouse}, service time rises again beyond the best-performing fleet-size range. Thus, over-provisioning agents can be counterproductive even though demand and workspace are fixed. The comparison supports a narrower conclusion: subsystem-local tokens reduce planning cost relative to a global token over the tested fleet sizes while retaining comparable service quality.

\subsection{Scope and limitations}
The evaluation covers two fixed workspaces and does not establish workspace-size or asymptotic scalability. The A* expansion term $S_i$ in Sec.~\ref{sec:controller} can grow substantially with map size, congestion, and reservation horizon. Sensitivity to $W$, $M_{\max}$, and $k_{\max}$ is not swept; all are held constant to isolate the interface mechanisms. Since $k_{\max}=8$ and every tested configuration has $|H|\leq 8$, the cap never binds, so the experiments do not measure the effect of dock-candidate truncation in larger interfaces. The two comparator studies use different independent variables because they isolate different mechanisms, and no unified three-way ranking is claimed. Longer or hierarchical subsystem networks, reverse or bidirectional flow, deadlines and priorities, execution uncertainty, broader parameter sweeps, and spatial scaling remain future work.
\section{Conclusion}
\label{sec:conclusion}
We introduce MS-MAPD-BHS, a lifelong pickup-and-delivery problem in which region-confined fleets exchange payloads through buffer-limited handover stations. HARR coordinates subsystem-local Token Passing planners through shared dock reservations and a deterministic rolling-horizon projection of committed buffer events. Under perfect execution, accepted dock operations are collision-free and buffer-safe within the enforced horizon, although the method is explicitly incomplete. In simulation, \textsc{HARR} achieves up to 77\% higher throughput and 92\% lower backlog than \textsc{FixedDock} at moderate load and reduces planning time relative to \textsc{MonoSBTP} at comparable service time.

Future work includes implementing and evaluating longer and hierarchical subsystem chains, incorporating deadlines and priority-aware assignment, handling execution uncertainty and bidirectional transfers, and studying workspace-size, parameter-sensitivity, and stability scaling.
\bibliographystyle{IEEEtran}
\bibliography{reference_small}
\end{document}